%% file: main.tex
\documentclass[runninghead,a4paper]{llncs}
\usepackage{makecell}
\usepackage{hyperref}
\usepackage{graphicx}
\usepackage{xspace}
\usepackage{array}
\usepackage{multirow}
\usepackage{makecell}
\usepackage{xr}

\begin{document}

\author{Peter Pere\v{s}\'ini \and Vladim\'ir Bo\v{z}a \and Bro\v{n}a Brejov\'a \and Tom\'a\v{s} Vina\v{r}}
\institute{
  Faculty of Mathematics, Physics and Informatics, Comenius University in Bratislava, Mlynsk\'a dolina, 842 48 Bratislava, Slovakia \\
\email{peter.peresini@fmph.uniba.sk}
}
\date{}

\input{content.tex}

\bibliography{main}
\bibliographystyle{splncs04}

\input{scontent.tex}

\end{document}

%% file: content.tex
\title{Nanopore Base Calling on the Edge}
  \maketitle
  
\begin{abstract}

We developed a new base caller DeepNano-coral for nanopore sequencing, which is optimized to run on the Coral Edge Tensor Processing Unit, a small USB-attached hardware accelerator.
  To achieve this goal, we have designed new versions of two key components used in convolutional neural networks for speech recognition and base calling.
  In our components, we propose a new way of factorization of a full convolution into smaller operations, which decreases memory access operations, memory access being a bottleneck on this device.
  DeepNano-coral achieves real-time base calling during sequencing with the accuracy slightly better than the fast mode of the Guppy base caller and is extremely energy efficient, using only 10W of power.

Availability: \href{https://github.com/fmfi-compbio/coral-basecaller}{\detokenize{https://github.com/fmfi-compbio/coral-basecaller}}


\end{abstract}

\section{Introduction}

MinION by Oxford Nanopore Technologies (ONT) is a portable DNA sequencer that measures electric current as DNA passes through nanopores.
  Electrical signals produced by the device need to be translated into sequences by a base caller software.
  Base calling of nanopore reads is a non-trivial task, and existing tools require powerful hardware and high energy consumption to operate in real time.

In this paper, we present a new base caller DeepNano-coral, which runs on the Coral accelerator featuring the Edge tensor processing unit (TPU), a small, energy-efficient, cheap USB-connected device.
  DeepNano-coral can process approximately 1.5 million signals per second, which is enough to provide real-time base calling for a MinION device.
  This makes our base caller ideal for field sequencing applications, where power efficiency and low hardware requirements are highly desirable.
  Real-time base calling is also essential in unlocking some of the most promising MinION device capabilities, such as its ability to adapt the run length to the sample composition, or selective sequencing \cite{loose2016real}.

Current base callers are typically based on deep neural networks.
  Guppy, a base caller provided by ONT, is based on recurrent neural networks (RNN), and provides two different architectures: a fast base caller, which can base call with 85-92\% median read accuracy in real time, using recent GPU cards, and a high-accuracy base caller (90-96\% median read accuracy), which is too slow to be used in real time without specialized setup. DeepNano-blitz trades off a bit of accuracy in order to provide real-time base calling on a common CPU using a specifically engineered RNN, thus obviating the need for GPUs \cite{boza2020deepnano}.
  Other RNN-based base callers, including Chiron \cite{teng2018chiron}, are too slow for real-time base calling.
  Another class of nanopore base callers is based on convolutional neural networks (CNN). In particular, Bonito v.0.2 \cite{bonito2020} adapts Jasper/Quartznet \cite{kriman2020quartznet,li2019jasper} speech recognition architecture to base calling tasks.
  At the time of writing, Bonito provided the most accurate base calling, however, the time requirements exceed even the Guppy high-accuracy mode.

  The Coral Edge TPU accelerator by Google is a limited device, which was designed mostly for vision tasks, such as image classification \cite{coral}.
  It contains only 8 MB of memory (used for storing both model weights and intermediate tensors), it only works with 8-bit integers (while GPUs typically work with 32-bit floating point numbers), and the compiler and libraries provide only a limited set of building blocks, optimized mostly for CNNs with small receptive fields, which are typically used in image processing.
  Such a configuration mostly excludes possibility of adaptation of RNN-based architectures and even adapting CNN-based architectures, such as Bonito, is a challenge, due to large size of the network and the use of large receptive fields.

Our new base caller DeepNano-coral, running on the Edge TPU, provides real-time base calling that is significantly more energy efficient than existing approaches.
   To achieve this goal, we introduce the following innovations:

\begin{itemize}

  \item A novel component \textbf{$k$-blueprint-separable-convolution}, which replaces separable convolutions as a building block for CNNs.
    A separable convolution approximates a full convolution by using a depthwise operation and a pointwise operation, which are less computationally intensive. The $k$-blueprint-separable convolution factorizes the convolution into the two parts differently, in effect reducing the depthwise operation at the cost of increasing computation in the pointwise operation.
    Even though the new convolution component has a higher number of parameters and floating point operations (flops), it is more efficient on the Edge TPU, since in this architecture, depthwise operations do not fully utilize the hardware \cite{xiong2020mobiledets,gupta2020accelerator}, possibly due to being bound by the memory bandwidth.
  
  \item \textbf{A new design of the residual block}, which is a fundamental building block of the QuartzNet speech recognition architecture and was also deployed for base calling in Bonito.
    To improve its performance on the Edge TPU, we add a compression operation at the start of the residual block, taking $x$ consecutive data samples of $C$ channels each, and converting them into a single compressed sample of $Cy$ channels.
    Using compression ratio $x/y<1$, we save memory and allow subsequent convolutions to effectively mix $x$ original samples and thus increase the receptive field of the block.

  \item A surprising observation that \textbf{identity initialization of some parts of the architecture helps the training and improves the prediction accuracy} in some circumstances, which contradicts usual recommendations for initializing parameters of neural networks before training.
  
\end{itemize}

Our experiments show that DeepNano-coral achieves the accuracy comparable to other real-time base callers, Guppy fast and DeepNano-blitz.
  Such accuracy is sufficient for real-time monitoring tasks, such as monitoring barcode composition in pooled libraries or species composition in environmental or clinical samples \cite{boza2020deepnano}.
  DeepNano-coral achieves this goal much more energy efficiently, using only 0.6--0.7 Wh of energy to base call a test sample of 40.8 Mbp of nanopore sequences at a speed of 1.54 million signal samples per second (on the same setup, the closest competitor, Guppy fast, uses 1.4 Wh of energy, processes 4.37 million signal samples per second, with up to 2 percentage points lower accuracy depending on the data set).

\paragraph{Background.}

Nanopore base calling translates the electrical signals produced by the sequencer into a sequence of DNA bases.
  The signal level depends on the context of about 5--12 DNA bases passing through the nanopore.
  The signal is read about 4000 times per second and DNA moves through the pore at the speed of approximately 450 bases per second, but the speed is rather uneven.
  This means that on average each shift of the context by one base corresponds to roughly 9 measured values with a large variance.
  This makes the problem somewhat similar to speech recognition.
  Note that different contexts may produce very similar signal levels and that there is a significant amount of noise present in the signal readouts, complicating the base calling problem.

\begin{figure}[t]
    \centering
  
    \includegraphics[width=0.7\textwidth]{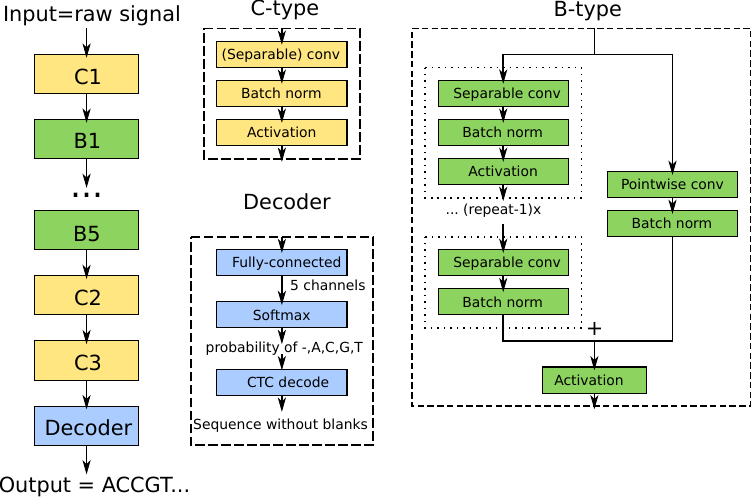}
    
    \caption{
      \label{fig:bonito_arch} Bonito architecture
    }
  \end{figure}

The work in this paper is based on the QuartzNet architecture \cite{kriman2020quartznet} for speech recognition, which has also been used in the Bonito base caller \cite{bonito2020} developed by ONT.
  Briefly, a window of the raw signal of length $T$ is used as an input to a deep CNN which uses several types of blocks to process the signal (see Figure \ref{fig:bonito_arch}).
  In the final decoder block, the network produces a tensor with 5 output channels.
  The five channels are converted by the softmax function into probability distributions over possible outputs A,C,G,T,- at each position, with dash corresponding to an empty output.
  Finally, the CTC layer \cite{hannun2017sequence} chooses a DNA sequence with the highest posterior probability.

In the QuartzNet / Bonito architecture, convolutions are organized into building blocks of two types B and C.
  The structure of a C-type block is simply a sequence of three layers: a convolutional layer, a batch normalization \cite{ioffe2015batch} (a layer that renormalizes channel values and stabilizes gradients for better training), and an activation function (Bonito uses Swish \cite{ramachandran2017searching}).

The B-type blocks use residual skip connections.
  The input signal is split into two branches.
  The main branch consists of $R$ copies of a C-type sub-block, with the last copy omitting the activation function.
  The second branch, called skip connection, consists of a pointwise convolution and batch normalization.
  The two branches are summed together and an activation is applied to the output.

The resulting network used in Bonito is large and computationally intensive.
  Some intermediate results reach size of up to $B\times T/3\times 464,$ where $B$ is the number of sequences combined to a batch, and $T$ is the length of the sequence.
  The network has 36 convolutional layers with 6.6 million parameters in total, requiring roughly 2.2 million multiplications per sample.

\section{Methods}

In this section, we present the architecture of our new base caller designed for the Edge TPU. Our architecture is inspired by the Bonito CNN, which was drastically scaled down and key components were replaced by the enhancements described here. Further technical details regarding adapting Bonito-like architecture to the Edge TPU are described in the Supplement.

\paragraph{The $k$-blueprint-separable convolutions.}

A convolutional layer is the dominant building block of many neural network architectures, mostly in the domain of image recognition, but recently also for automated speech recognition \cite{li2019jasper,kriman2020quartznet,oord2016wavenet}.
  In this paper, we consider 1D convolutions, which take as an input a tensor $X$ of dimensions $(T,C_{in})$ representing a data stream of length $T$, each data point containing $C_{in}$ values called \emph{channels}.
  To apply a convolution with odd depth $D$, the input tensor $X$ is first padded with $\lfloor\frac{D}{2}\rfloor$ zeros at the beginning and at the end.
  Then the output tensor $Y$ with dimensions $(T,\ C_{out})$ is computed as follows: $Y_{t,j}=\sum_{0\leq d<D,{0\leq i}<C_{in}}X_{t+d,i}W_{j,d,i}+B_{j},$ where $W$ and $B$ are trained weights representing convolution kernel weights and bias.

\begin{figure}[b]
    \centering
  
    \includegraphics[width=0.6\textwidth]{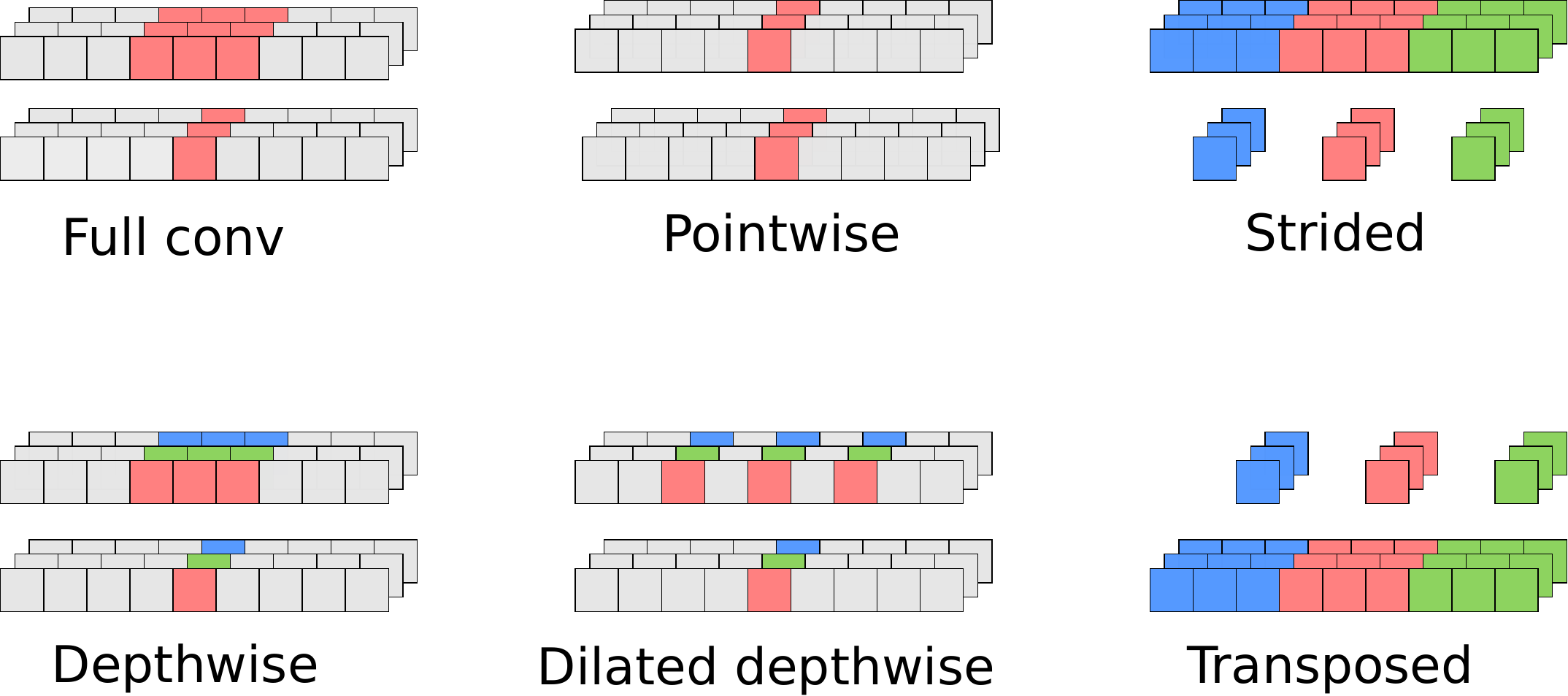}
    
    \caption{
      \label{fig:receptive} Receptive fields for basic types of convolutions}
    
  \end{figure}

An obvious drawback of full convolutions is a large number of parameters ($C_{out}D{C}_{in}$) and required flops ($TC_{out}DC_{in}$).
  A standard solution is to use a \emph{separable convolution} \cite{mamalet2012simplifying}, which is an approximation of the full convolution by a composition of two operations: depthwise and pointwise.
  The depthwise operation works on each channel separately: $Z_{t,j}=\sum_{0\leq d<D}X_{t+d,j}W^{(D)}_{d,j}+B^{(D)}_{j}$.
  This is followed by the pointwise operation, which mixes the channels at each time point: $Y_{t,j}=\sum_{0\leq i<C_{in}}Z_{t,i}W^{(P)}_{j,i}+B^{(P)}_j$.
  This reduces the flops from  $TC_{out}DC_{in}$ to $T(DC_{in}+C_{out}C_{in})$.
The ordering of pointwise and depthwise operations was chosen somewhat arbitrarily, and reversing it may improve the accuracy \cite{haase2020rethinking}.
  The variant with the reversed order is called a \emph{blueprint-separable convolution}.
  Figure \ref{fig:receptive} illustrates receptive fields of basic operations used in convolutions.

Recent works \cite{xiong2020mobiledets,gupta2020accelerator,dwsep2020} indicate that separable convolutions do not always improve the speed on non-CPU architectures, because the depthwise operation requires a smaller ratio of flops to memory operations, which are generally slow.
  A full convolution with depth $D=3$ can be faster than a separable convolution with the same depth \cite{xiong2020mobiledets}.
  Full convolutions with small depths are thus feasible in image recognition, while in base calling, the kernels need to be much larger.

\begin{figure}[b]
    \centering
  
    \includegraphics[width=0.9\textwidth]{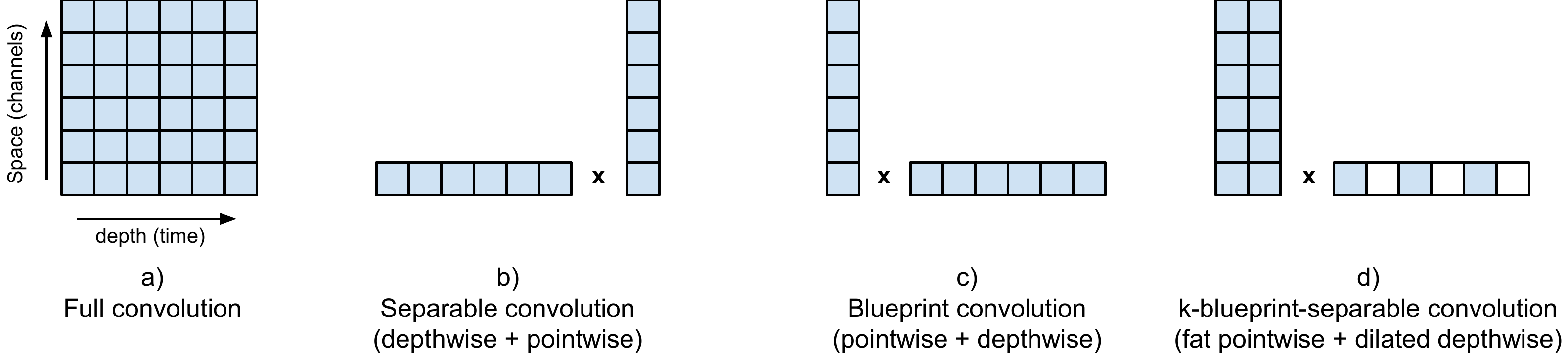}
    
    \caption{
      \label{fig:conv_comparison} Comparison of different convolution factorizations
    }
  \end{figure}

Our design of $k$-separable convolutions is heavily influenced by this observation.
  Our goal is to reduce the time-consuming depthwise operations using dilation with step size $k$, and compensate by replacing the pointwise operation by a convolution operating on a window of size $k$ instead of a single point, as illustrated in Figure~\ref{fig:conv_comparison}.

Namely, we start with what we call a \emph{fat-pointwise} operation,
which is a standard convolution of depth $k$:
 $Z_{t,j}=\allowbreak\sum_{0\leq d<k,0\leq i<C_{in}}X_{t+d,i}W^{(P)}_{j,d,i}+B^{(P)}_j$.
The second step uses a dilated depthwise operation with depth $D/k$, which skips points by using dilation $k$:
 $Y_{t,j}=\sum_{0\leq d<D/k}Z_{t+dk,j}W^{(D)}_{d,j}+B^{(D)}_j$.
This reduces the depthwise kernel (and thus memory I/O) by a factor of $k$, while retaining the receptive field $D$ of the whole layer.

Note that the special case of $k=1$ leads to a standard blueprint convolution, while we typically use $k=3$, which on the Edge TPU roughly maintains the same computation time as separable convolutions, while increasing the accuracy.

\begin{figure}[b]
    \centering
 
    \begin{minipage}[t]{1.0\textwidth}
      \centering
    \includegraphics[width=0.45\textwidth]{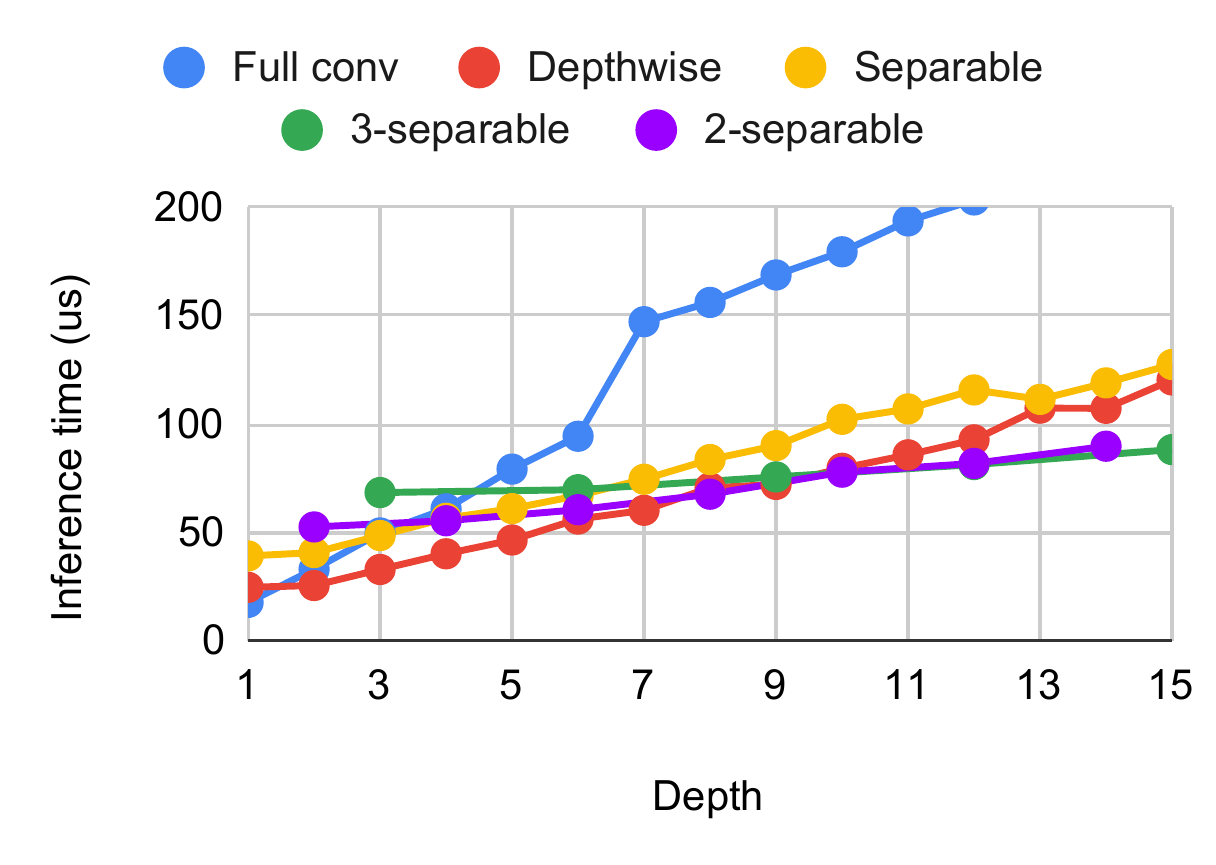}
    \includegraphics[width=0.45\textwidth]{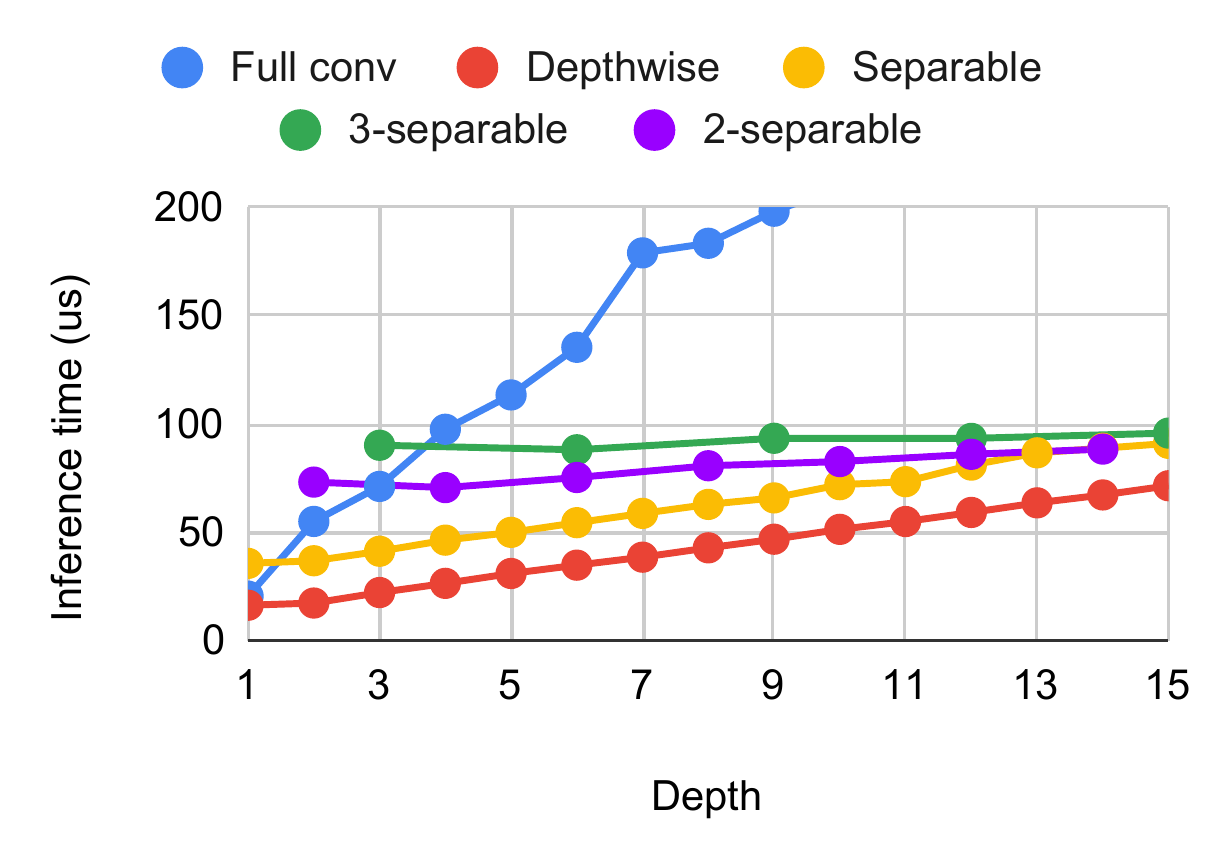}
    \end{minipage}

    \caption{
       \label{fig:microbenchmark} Inference time of different convolutions for a) $C_{in}=C_{out}=128$ and tensor size $(4,1668,128)$.
        b) $C_{in}=C_{out}=256$ and tensor size $(4,556,256)$.
          Note that pointwise corresponds to a full convolution with depth 1.
    }
  \end{figure}

Figure \ref{fig:microbenchmark} demonstrates the performance of $k$-separable convolutions on two configurations used in our experiments in the next section.
  Our $k$-separable convolutions offer running times comparable to separable convolutions, while providing roughly $k$ times more parameters, which increases their expressive power.

\paragraph{Residual block with depth-to-space compression.}

  \begin{figure}[t]
    \centering
  
    \includegraphics[width=1.0\textwidth]{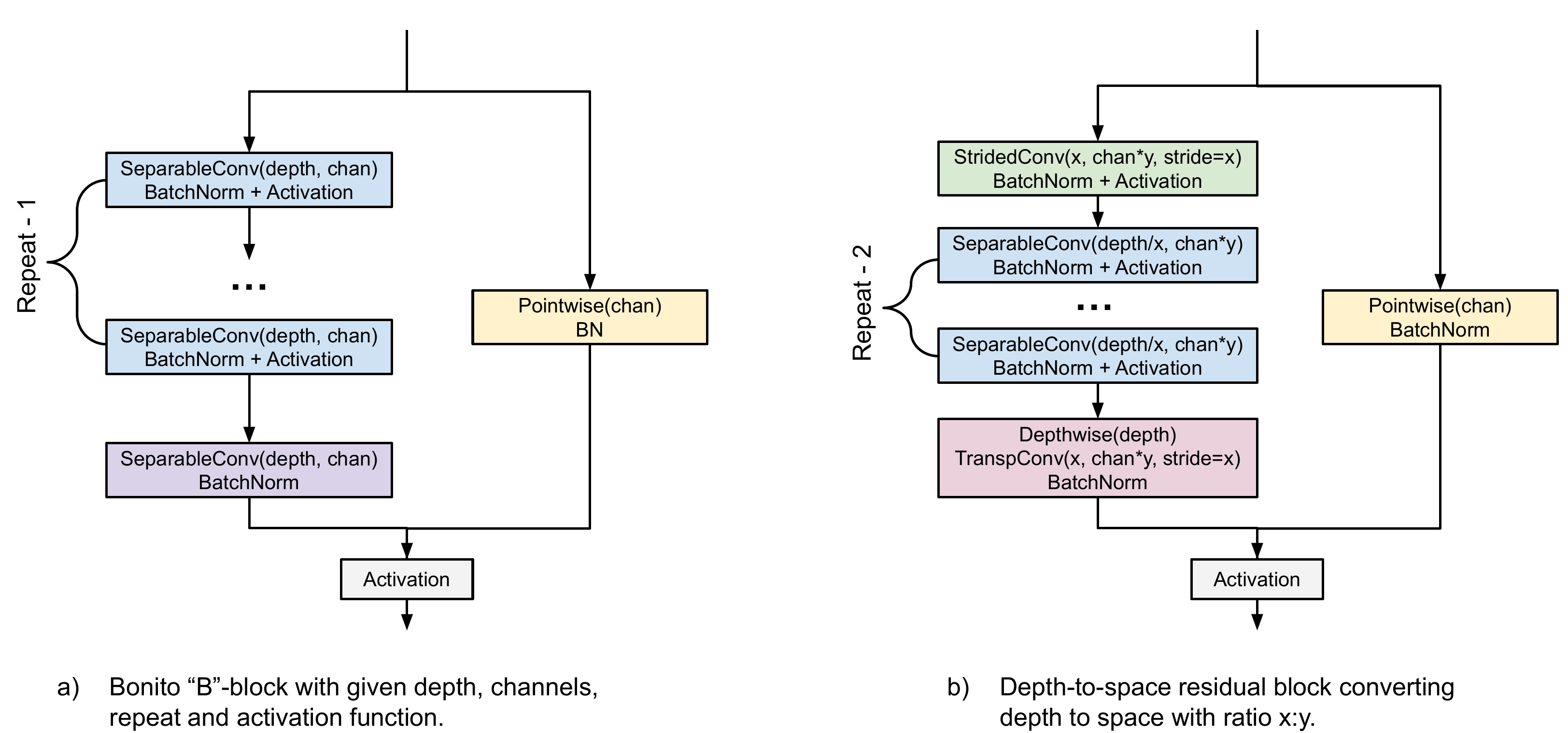}
    
    \caption{
      \label{fig:d2s_block} Residual block with depth-to-space compression
    }
  \end{figure}

Our second change also targets reduction of the depthwise convolution.
  As shown in Figure \ref{fig:microbenchmark}, when we apply convolutions on a shorter input, we can use more channels in a comparable time.
  Our idea is to redesign the residual block of the CNN (B-type block in Figure~\ref{fig:bonito_arch}) so that we compress its depth and increase the number of channels.
  In particular, compression with depth-to-space ratio $x:y$ means converting input tensor $(T,C)$ to tensor $(T/x,Cy)$ using a strided convolution with both depth and stride set to $x$ (see Figure~\ref{fig:receptive}).
  This convolution takes $x$ consecutive data samples of $C$ channels and converts them into a single compressed sample of $Cy$ channels.
  At the end of the residual block, we restore the original dimensions with a strided transposed convolution.
  This makes the new block a drop-in replacement for the original B-type block design (see Figure~\ref{fig:d2s_block}).

Compression ratio $x/y<1$ saves memory, which is essential due to limited Coral resources.
 While compression may sometimes decrease accuracy, the network may learn to de-duplicate information from consecutive data samples, and thus prevent data loss.
  In fact, any subsequent pointwise operations effectively operate on $x$ original samples, yielding increased receptive fields.
  Thus, we can further lower the depth of the depthwise operation in the block, offsetting  larger computation of pointwise operations, which were increased by a factor of  $y^2/x$.
  In our experiments, compression ratio 3:2 works well on Coral.

To complete the residual block, we add the depthwise operation before the decompression.
  While the original B-type block repeats separable convolutions $R$ times, we repeat them $R-2$ times, since we consider the compression and decompression blocks as replacements for two separable convolutions.

\paragraph{Identity initialization.}

A proper neural network initialization can affect both trainability and final accuracy of models \cite{glorot2010understanding,zhang2019fixup,sutskever2013importance,le2015simple,mishkin2015all}.
  A standard way of initializing CNN architectures is to draw the entries of weight matrices from the uniform distribution $U(-k,k)$, where $k=6/\sqrt{C_{in}+C_{out}}$, and to set the bias terms to zero \cite{glorot2010understanding}.
  The weighting factor $k$ is used to keep the gradients from vanishing or exploding as the number of layers increases.
  Recent introduction of BatchNorm however obviates such problems, as the results are renormalized \cite{ioffe2015batch}.

In some cases, task specific initialization may bring an improvement over the generic initialization strategies \cite{le2015simple}, and this proved to be the case for our base calling application as well.
  We initialize all $k$-separable blocks within the compressed main branch to near-identity, that is, depthwise kernels are initialized as $W^{(D)}_{d,j}=\delta_{\lfloor (D/K)/2 \rfloor,d}$ and fat pointwise kernels to $W^{(P)}_{j,d,i}\sim \delta_{\lfloor K/2 \rfloor,d}(\delta_{i,j}+U(-\epsilon,\epsilon)),$ where $\delta_{x,y}=1$ if and only if $x=y$.
  We experimented with several other initializations and observed that setting the depthwise operations to identity helps the most, while setting pointwise operations to identity brings only a small additional improvement.
  On the other hand, initialization of the skip connection as well as of the compression/decompression block does not seem to affect the results significantly.

In our experiments, the identity initialization described above speeds up the process of training and decreases overfitting (see Results).
  We believe that this surprising effect is explained by the properties of the base calling task.

Due to the nature of nanopore raw sequencing data, base calling is composed of two tasks.
  First, the input signal needs to be segmented into events, each event corresponding to the shift of the DNA currently read by the nanopore head by one base.
  The length distribution of these events is highly variable.
  The second task is to recognize the base under the nanopore head given the context of several events.
  Although initially base callers have performed these tasks separately, modern neural network approaches combine them into a single optimization problem.

One would assume that the second task of correctly identifying bases is the core of the problem.
  A quick experiment in which Bonito is provided with an additional binary input indicating event boundaries (as determined from a ground-truth alignment to a reference) shows otherwise.
  In particular, the additional input dramatically speeds up the training so that the network can in minutes outperform days-long Bonito training.
  While the modified network cannot be used for practical base calling (because the base caller obviously cannot receive ground-truth event boundaries as an input), it suggests that identification of events is in fact the harder part of the base calling task.
  This is further corroborated by the fact that even a simple logistic regression can distinguish purines A,G from pyrimidines C, T in a correctly segmented signal.

Our depthwise identity initialization indeed makes sense assuming that the network spends much more time learning how to split the raw signal into events rather than recognizing individual bases.
  Identity initialization may allow the network to learn the easy task of distinguishing bases first and then spend the rest of its capacity on learning intricate time-dependencies without the need for unlearning spurious long-range correlations that may have been introduced by random initial weights.

\section{Results}

In this section, we compare the speed, energy consumption, and accuracy of DeepNano-coral with other tools
  on a data set of R9.4.1 reads from \emph{K. pneumoniae} \cite{wick2019performance} and human \cite{jain2018nanopore} (see Supplement).
  The base calls were mapped to the reference using minimap2 \cite{li2016minimap}.

DeepNano-coral slightly outperforms Guppy fast in most accuracy measures (Table \ref{tab:accuracy}).
  Guppy fast would currently be a method of choice for live base calling on a computer with a recent GPU card (compute capability 6.2, 4GB of memory).
  As demonstrated earlier \cite{boza2020deepnano}, even slightly lower accuracy of DeepNano-blitz is sufficient for run monitoring, such as barcode composition or metagenomic analysis.
  Note that DeepNano-blitz provides real-time base calling on a CPU without the use of any accelerator.
  Guppy in the high accuracy (hac) mode illustrates accuracy gains possible with more extensive computational resources typically beyond the possibilities of real-time base calling.

\begin{table}[t]
  \centering
  
  \caption{
    \label{tab:accuracy} Comparison of base calling accuracy.
      Read accuracy is computed as one minus the ratio of the alignment edit distance and the base call length. We report the median read accuracy.
  }

  \begin{tabular}{l@{\quad}c@{\quad}c@{\quad}|@{\quad}c@{\quad}c}
    \hline  		& \multicolumn{2}{c@{\quad}|@{\quad}}{\emph{Klebsiella pneumoniae}}		& \multicolumn{2}{c}{Human} \\
      Base caller		& Mapped		& Median accuracy		& Mapped		& Median accuracy \\
    \hline
   Guppy 4.0.11 hac		& 100\%		& 94.7\%		& 93.4\%		& 89.8\% \\
   Guppy 4.0.11 fast		& 100\%		& 91.2\%		& 92.1\%		& 84.7\% \\
   DeepNano-coral		& 100\%		& 92.4\%		& 88.5\%		& 87.2\% \\
   DeepNano-blitz 80		& 100\%		& 90.4\%		& 86.1\%		& 84.3\% \\
    \hline
  \end{tabular}
\end{table}

We have measured the speed and energy consumption on two computers with different setups (Table \ref{tab:energy}),
  a desktop (i7-7700k 4 core CPU; NVIDIA GTX 1650 GPU) and a laptop (i7-7700HQ 4 core CPU) incapable of running the GPU version of Guppy.
  To run DeepNano-coral, we have attached the Coral Edge TPU device through USB 3.0 interface.

On both computers, DeepNano-coral achieved the speed necessary for live base calling (1.5M signals per second) and used less than 11W (computed as a difference between the idle energy consumption and the consumption during base calling).
  On our testing set, the total energy spent on base calling was 0.58-0.68Wh, roughly half of the energy used by Guppy fast on the desktop.
  Although Guppy fast consumed less energy when baseline is included due to its shorter running time, in a practical setting, this would not translate to energy savings as the computer needs to run throughout the sequencing.

DeepNano-coral runs on GPU at lower speed and with higher energy consumption than GPU- and CPU-optimized software. This underlines the importance of optimization of the network architecture for a particular platform.

\begin{table}[t]
  \centering
  
  \caption{
    \label{tab:energy} Energy consumption and speed of different base callers (DN=DeepNano)
  }

  \begin{tabular}{lccccc}
    \hline  Base caller		& \makecell{Power\\(W)}		& \makecell{Speed\\(signals/s)}		& \makecell{Time\\(s)}		& \makecell{Energy for\\base calling (Wh)}		& \makecell{Total energy\\(Wh)} \\
    \hline  {\bf Desktop} & 		& 		& 		& 		&  \\
    Idle baseline 		& 62		& -		& -		& -		& - \\
    DN-coral		& 72-73		& 1.52 M		& 234		& 0.68		& 4.71 \\
    DN-blitz 80 (4 threads)		& 168-170		& 2.12 M		& 168		& 4.94		& 7.84 \\
    DN-blitz 80 (2 threads)		& 120-122		& 1.13 M		& 316		& 5.09		& 10.53 \\
    Guppy fast		& 110		& 3.32 M		& 107		& 1.42		& 3.27 \\
    Guppy hac		& 135		& 79k		& 4495		& 91.14		& 168.56 \\
    DN-coral on GPU		& 154		& 1.34M		& 265		& 6.8		& 11.3 \\
    {\bf Laptop}	& 		& 		& 		& 		&  \\
    Idle baseline		& 18		& -		& -		& -		& - \\
    DN-coral		& 27		& 1.51M		& 235		& 0.58		& 1.76 \\
    DN-blitz 80 (4 threads)		& 73		& 1.53M		& 232		& 3.54		& 4.70 \\
    DN-blitz 80 (2 threads)		& 56		& 907k		& 392		& 4.13		& 6.09 \\
    \hline
  \end{tabular}
\end{table}

To further illustrate the impact of our new network designs on the base calling accuracy, we started with the small Bonito architecture (see the Supplement), in which we replaced various components by our new designs presented in the Methods section.
   In these experiments, we modify only B-type (residual) blocks, keeping the standalone C-type blocks the same.
  We however verified that altering configuration of these C-type blocks does not affect the accuracy significantly.

\begin{figure}[t]
    \centering
  
    \includegraphics[width=0.9\textwidth]{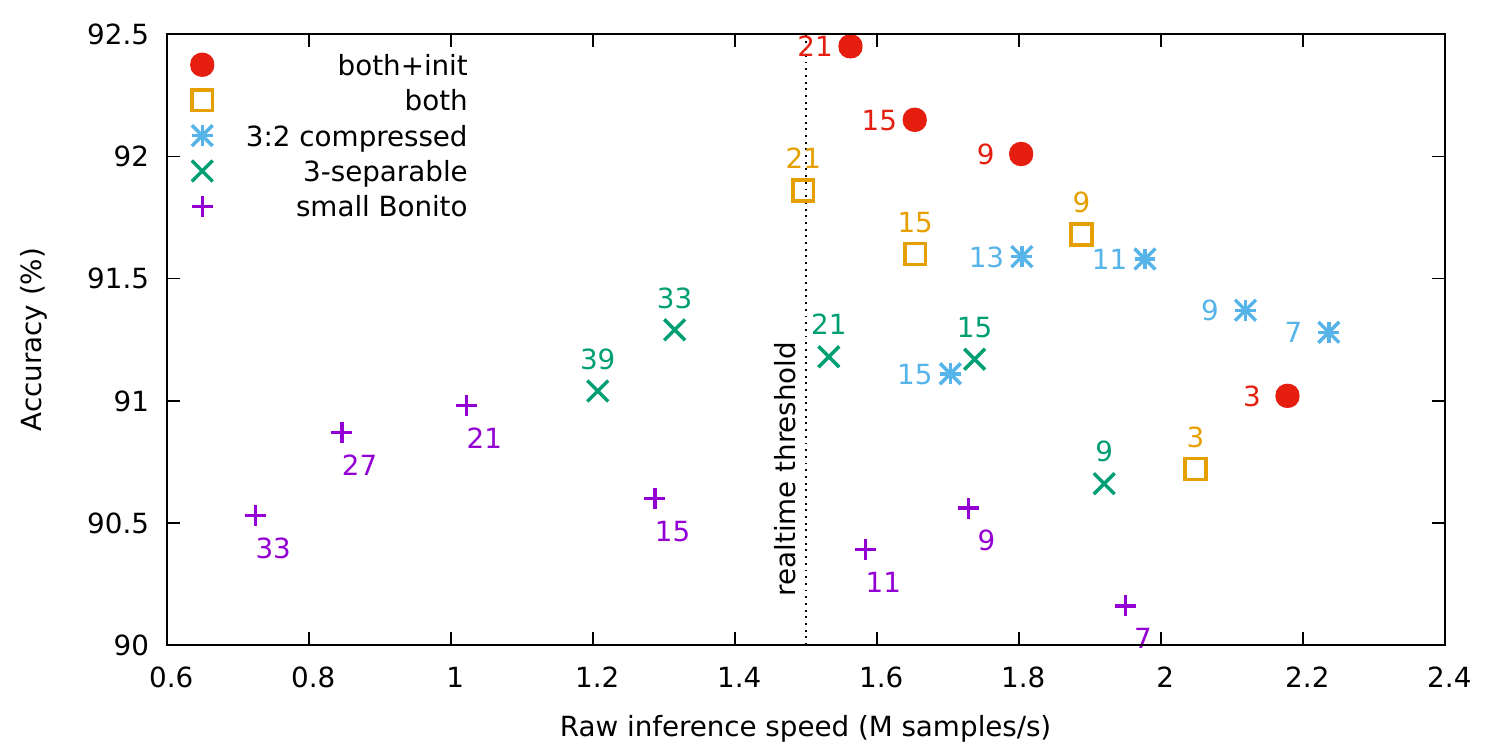}
    
    \caption{
       \label{fig:ablation} Speed-vs-accuracy frontier for various depthwise kernel sizes
    }
  \end{figure}

Figure \ref{fig:ablation} shows the accuracy and speed starting with the small Bonito and adding the following features: 3-blueprint-separable convolutions, compression with ratio 3:2, combination of the two, and finally the identity initialization.
  For each variant, we test several kernel depths.
  Note that 3-separable convolutions have a symmetrical receptive field only for depth of size $k=3(2n+1)$.
  In most experiments, we stop at kernel size 21, because larger kernels lead to base calling speed below the speed of sequencing.
  In general, adding our modifications increases the accuracy at comparable speed, and the most accurate version is the one with all our improvements combined.


\section{Conclusions and future work}

We have presented a base caller capable of running on the Edge TPU in real time.
  To do so, we have designed new types of blocks which can be used as drop-in replacements for separable convolutions and QuartzNet-style residual blocks, potentially improving their speed/accuracy tradeoff in other applications as well.

From a practical standpoint, our work enables real-time base calling with low energy consumption on modest hardware with addition of a \$70 USB device.
  This contribution will help researchers attempting nanopore sequencing in field conditions with limited energy resources.
  Using Edge TPU as an alternative to GPU chips may also help to design new devices specifically targeted at nanopore sequencing, analogous to the MK1C device manufactured by ONT.

Further research into decreasing the size of the base calling neural networks may yield even better results on small accelerators.
  One option is to use knowledge distillation \cite{hinton2015distilling}, where the smaller network is trained on outputs from a larger network.
  Another avenue is to consider a richer set of outputs from the network.
  In our case, the softmax layer output probabilities over the $\{A,C,G,T,-\}$ alphabet, which is followed by CTC decoding.
  Guppy and Bonito v0.3 use a more complicated scheme, which could be adapted.
  The risk here is that we would need to do more intensive decoding on the CPU, which may become a bottleneck.

\paragraph{Acknowledgements.}

This research was supported in part by grants from the Slovak Research Grant Agency VEGA 1/0458/18 (TV) and 1/0463/20 (BB), by the grant from Slovak Research and Development Agency APVV-18-0239, and by funding from the European Union's Horizon 2020 research and innovation programme under grant agreement No 872539 (PANGAIA).
  The work has also been supported by Google Cloud (VB).
  We gratefully acknowledge the support of NVIDIA Corporation with the donation of hardware used in this research.

%% file: scontent.tex
\renewcommand\thetable{S\arabic{table}}
\renewcommand\thefigure{S\arabic{figure}}
\renewcommand\thesection{S\arabic{section}}

\title{
Nanopore Base Calling on the Edge\\
Supplementary Material
}
\author{Peter Pere\v{s}\'ini \and Vladim\'ir Bo\v{z}a \and Bro\v{n}a Brejov\'a \and Tom\'a\v{s} Vina\v{r}}
\institute{
  Faculty of Mathematics, Physics and Informatics, Comenius University in Bratislava, Mlynsk\'a dolina, 842 48 Bratislava, Slovakia \\
\email{peter.peresini@fmph.uniba.sk}
}
\date{}
\maketitle

\section{Engineering neural networks for Coral Edge TPU}

In this section, we discuss practical steps needed to adapt the Bonito GPU-tuned base caller architecture to run on and fully utilize Edge TPU accelerator.
  These observations can be useful also to others who want to optimize different neural network architectures for this platform.

Our first step was to create a scaled-down version of Bonito, capable of running on the Edge TPU. Bonito uses large convolutions, with up to 464 output channels.
  Our experiments suggest that the performance of the Edge TPU accelerator severely deteriorates beyond approximately 128 channels.
  To stay safely within these bounds, we decrease the maximum number of channels to 128.
  We also scale down the depth of the depthwise operation, considering the range of 9-33, as the performance cost of larger depthwise kernels is noticeable (see the main text).
  The final version of DeepNano-coral uses kernel depth 21.

To avoid an extensive architecture search, we use a more uniform configuration of building blocks.
  In particular, all residual B-type blocks use five convolution blocks, which is a middle ground compared to the original Bonito configuration.

Another issue is related to the quantization.
  It would be ideal to fuse the activation function to the preceding convolution layer, but the development tool chain does not support this. It also does not support the Swish \cite{ramachandran2017searching} activation function used in Bonito. For these reasons, we have used ReLU6 as the activation function, which can be fused into preceding convolution layers by a simple value clipping of int32 accumulator values used in matrix multiplication.

Further issues are related to limitations of the available development tools for the Edge platform.
  Typically, neural network inference is done in batches, with several inputs processed simultaneously to optimally utilize hardware capacity.
  However, this setup is not supported in the current development tool chain.
  Another problem is that 1D convolutions are internally converted to 2D convolutions, which adds a reshape operation before and after the convolution.
  This has a severe performance impact, since reshape operations are memory intensive.

To solve both these problems at the same time, we transform multiple 1D inputs into a single 2D ``image'' of dimensions $B\times T$, where $B$ is the batch size and $T$ is the sequence length.
  The whole network is then rewritten to an equivalent network operating on this ``image'' using 2D convolutions.
  In our network, considering the utilization of the device and the target speed, we use $B=4$ and $T=5004$ ($T$ must be a multiple of 9).
  Note that splitting the input signal into slightly overlapping chunks of 5004 observations is a reasonable compromise between overhead imposed by the overlaps and the capacity of the device.

\begin{table}[t]
  \centering
  \caption{
    \label{tab:baseline_config} Baseline small Bonito architecture we use in our experiments.
  }
  \begin{tabular}{|c|c|}
    \hline  C1		& Conv(filters=128, depth=9, stride=3) + BatchNorm + ReLU6 \\
    \hline  B1-B5		& Residual with 5x SeparableConv(filters=128, depth=~7-33)  \\
    \hline  C2		& SeparableConv(filters=128, depth=11) + BatchNorm + ReLU6 \\
    \hline  C3		& Conv(filters=64, depth=7) + BatchNorm + ReLU6 \\
    \hline  Decoder		& Pointwise(filters=5) + Softmax \\
    \hline
  \end{tabular}
  
\end{table}

After these modifications, we obtain a small Bonito-like architecture which the Edge TPU compiler is able to fit on the device.
  The overall architecture configuration is summarized in Table~\ref{tab:baseline_config}. Finally, the last softmax layer as well as CTC decoding are performed directly on the CPU.

\section{Training and Testing Sets}
\label{sec:arch_train_test}

We have used the combination of the following public data sets to train the models:

\begin{itemize}

  \item \emph{Taiyaki set:}
    Data set of 50k (downsampled to 5k) R9.4.1 reads from a PCR amplified DNA of \emph{E. coli} (SCS110), \emph{H. sapiens} (NA12878), and \emph{S. cerevisiae} (NCYC1052),
    published by Oxford Nanopore as a part of Taiyaki software
    (\href{https://github.com/nanoporetech/taiyaki/}{\detokenize{https://github.com/nanoporetech/taiyaki/}}).
  
  \item \emph{E. coli set:} A sample of 2804 ultra-long native R9.4.1 reads of \emph{E. coli }(MG1655) from Loman Lab
    (\href{https://lab.loman.net/2017/03/09/ultrareads-for-nanopore/}{\detokenize{https://lab.loman.net/2017/03/09/ultrareads-for-nanopore/}}).
  
  \item \emph{Human training:} A sample of reads from chr1 and chr2 of native R9.4.1 reads (except flowcell FAB49164)
    from the human reference standard CEPH1463 from nanopore whole human genome sequencing project \cite{jain2018nanopore}.
  
\end{itemize}

For testing, the following data sets were used:

\begin{itemize}

  \item \emph{Klebsiella}: a benchmark set of native R9.4 \emph{K. pneumoniae }reads \cite{wick2019performance}. We only used reads before the sequencing restart.
  
  \item \emph{Human testing:} a sample of native R9.4.1 reads from chr14, chr15, chr16 (flowcell FAB49164)
    from human reference standard CEPH1463 from nanopore whole human genome sequencing project \cite{jain2018nanopore}
  
\end{itemize}

The basic characteristics of all data sets are shown in Table~\ref{tab:datasets}.

\newcolumntype{R}[1]{>{\raggedleft\let\newline\\\arraybackslash\hspace{0pt}}m{#1}}
\newcolumntype{C}[1]{>{\centering\let\newline\\\arraybackslash\hspace{0pt}}m{#1}}

\begin{table*}[!ht]
  \centering

  \makebox[\linewidth]{
  \newcommand{\fix}{\phantom{xx}}
  \newcommand{\fixx}{\phantom{xxxx}}
  \begin{tabular}{|c|r|r|r|r|}
    \hline
      \textbf{Data Set} &
      \textbf{\# reads} &
      \multicolumn{1}{R{2.2cm}|}{\textbf{Total length \newline (in bp)}}        &
      \multicolumn{1}{R{2.5cm}|}{\textbf{Mean read length (in bp)}}     &
      \multicolumn{1}{R{2.5cm}|}{\textbf{Median read length (in bp)}} \\
    \hline  Taiyaki set                 & 5000\fix              & 28.8 Mbp\fix          & 5769\fixx             & 5459\fixx \\
    \hline  E. coli training set        & 2804\fix              & 90.6 Mbp\fix          & 32319\fixx            & 21758\fixx \\
    \hline  Human training set          & 1323\fix              & 11.5 Mbp\fix          & 8705\fixx             & 7111\fixx \\
    \hline  Human testing set           & 305\fix               & 2.7 Mbp\fix           & 8936\fixx             & 6231\fixx \\
    \hline  Klebsiella                  & 1788\fix              & 40.8 Mbp\fix          & 22613\fixx            & 17547\fixx \\
    \hline
  \end{tabular}
  }

  \caption{
     \label{tab:datasets} Overview of training and testing sets used in the study.
  }
\end{table*}

\section{Training procedure details}

We wrote our training pipeline in TensorFlow and the code is available at \url{https://github.com/fmfi-compbio/coral-training}.
Our training schedule uses 6000 miniepochs, one miniepoch being 15 optimizer steps on batch of 100 sequences of length 5004 (5004 being the nearest multiple of 9 targeting our desired sequence length 5000).
We use Adam optimizer with a schedule that uses rather high learning rate decaying linearly $LR(t) = 0.01 * (1-t)$ updated after each miniepoch, where $0 \le t \le 1$ denotes progress.
We have a short ramp-up period over first 10 miniepochs where we increase learning rate from 0 to the maximum value to avoid "shocking" the model with high learning rate from the start.
At the end of training we use standard TensorFlow post-training quantization to convert the model into a format compatible with Edge TPU compiler.
